\documentclass{article} 
\usepackage{iclr2025_conference,times}

\definecolor{cornflowerblue}{RGB}{100, 149, 237}

\usepackage{xcolor}         

\usepackage[utf8]{inputenc} 
\usepackage[T1]{fontenc}    
\definecolor{citecolor}{HTML}{0071BC}
\usepackage[pagebackref=false, breaklinks=true, letterpaper=true, colorlinks, citecolor=citecolor, bookmarks=false]{hyperref}
\usepackage{url}
\usepackage{multicol}
\usepackage{multirow}
\usepackage{url}
\usepackage{graphicx}
\usepackage{bm}
\usepackage{amsmath}
\usepackage{amssymb}
\usepackage{mathtools}
\usepackage{amsthm}
\usepackage{subcaption}
\usepackage{xspace}
\usepackage{bbding}
\usepackage{pifont}
\usepackage{colortbl}
\usepackage{enumitem}
\usepackage{overpic}
\usepackage{booktabs}
\usepackage[table]{xcolor}
\usepackage{dsfont}
\usepackage{wrapfig}
\makeatletter
\DeclareRobustCommand\onedot{\futurelet\@let@token\@onedot}
\def\@onedot{\ifx\@let@token.\else.\null\fi\xspace}
\def\eg{\emph{e.g}\onedot} 
\def\ie{\emph{i.e}\onedot} 
 
\def\etc{\emph{etc}\onedot}

\newcommand\figcaption{\def\@captype{figure}\caption} 
\newcommand\tabcaption{\def\@captype{table}\caption} 
\makeatother
\usepackage{algorithm}  
\usepackage{algorithmic}

\title{A Simple Baseline for Unifying Understanding, Generation, and Editing via Vanilla Next-token Prediction}

\author{Jie Zhu$^{1,2}$, Hanghang Ma$^4$, Jia Wang$^3$, Yayong Guan$^4$, Yanbing Zeng$^4$, Lishuai Gao$^4$, \\   \textbf{Junqiang Wu$^4$,  Jie Hu$^{4*}$, Leye Wang$^{1,2}$\thanks{Corresponding author}} \\
$^1$Key Lab of High Confidence Software Technologies (Peking University), Ministry of Education, China \\
$^2$School of Computer Science, Peking University, Beijing, China\\
$^3$School of Computer Science and Technology, University of Chinese Academy of Sciences, Beijing, China\\
$^4$Meituan\\
\texttt{zhujie@stu.pku.edu.cn, wangj.infinite@gmail.com, leyewang@pku.edu.cn} \\
\texttt{\{mahanghang, gaolishuai, zengyanbing02, lichen129,  hujie39\}@meituan.com}
}

\iclrfinalcopy 
\begin{document}

\maketitle

\begin{abstract}
In this work, we introduce \textit{Wallaroo}, a simple autoregressive baseline that leverages next-token prediction to unify multi-modal understanding, image generation, and editing at the same time. Moreover, Wallaroo supports multi-resolution image input and output, as well as bilingual support for both Chinese and English. We decouple the visual encoding into separate pathways and apply a four-stage training strategy to reshape the model's capabilities. Experiments are conducted on various benchmarks where Wallaroo produces competitive performance or exceeds other unified models, suggesting the great potential of autoregressive models in unifying multi-modality understanding and generation. Our code is available at \url{https://github.com/JiePKU/Wallaroo}.
\end{abstract}

\section{Introduction}

With the development of multi-modal understanding~\citep{liu2023visual, chen2024internvl, team2025kimi, wang2024qwen2} and visual generation~\citep{rombach2022high, zhu2024mole, li2024hunyuan, hong2022cogvideo, esser2024scaling, zhuunveiling, zhu2025auditing}, unifying understanding and generation has become a hot trend, as a key step toward the promising vision of artificial general intelligence. As a result, various efforts are devoted to this realm. Current methods~\citep{pan2025transfer, wu2025omnigen2, chen2025blip3, wang2025ovis, kou2024orthus, lin2025uniworld, zhou2024transfusion, ma2025janusflow, deng2025emerging, team2024chameleon, qu2025tokenflow, ma2025unitok, wu2025janus, chen2025janus, xie2024show, liao2025mogao, wang2024emu3, li2025onecat} could be roughly categorized into three classes. The first class views multi-modal understanding models as enhanced conditional encoders for following diffusion generation like OmniGen2~\citep{wu2025omnigen2}, leading to \textit{unidirectional} information interaction, \ie, from understanding to generation. The second class integrates auto-regressive understanding and diffusion generation within transformers such as Bagel~\cite{deng2025emerging}. However, the presence of diffusion noise in the representation leads to relatively low information interaction efficiency. The third class employs an autoregressive model with next-token prediction to understanding and generation, substantially reducing structural and training complexity while improving information interaction efficiency.

\begin{figure}[t]
\vskip -0.25in
\centering{\includegraphics[width=1\linewidth]{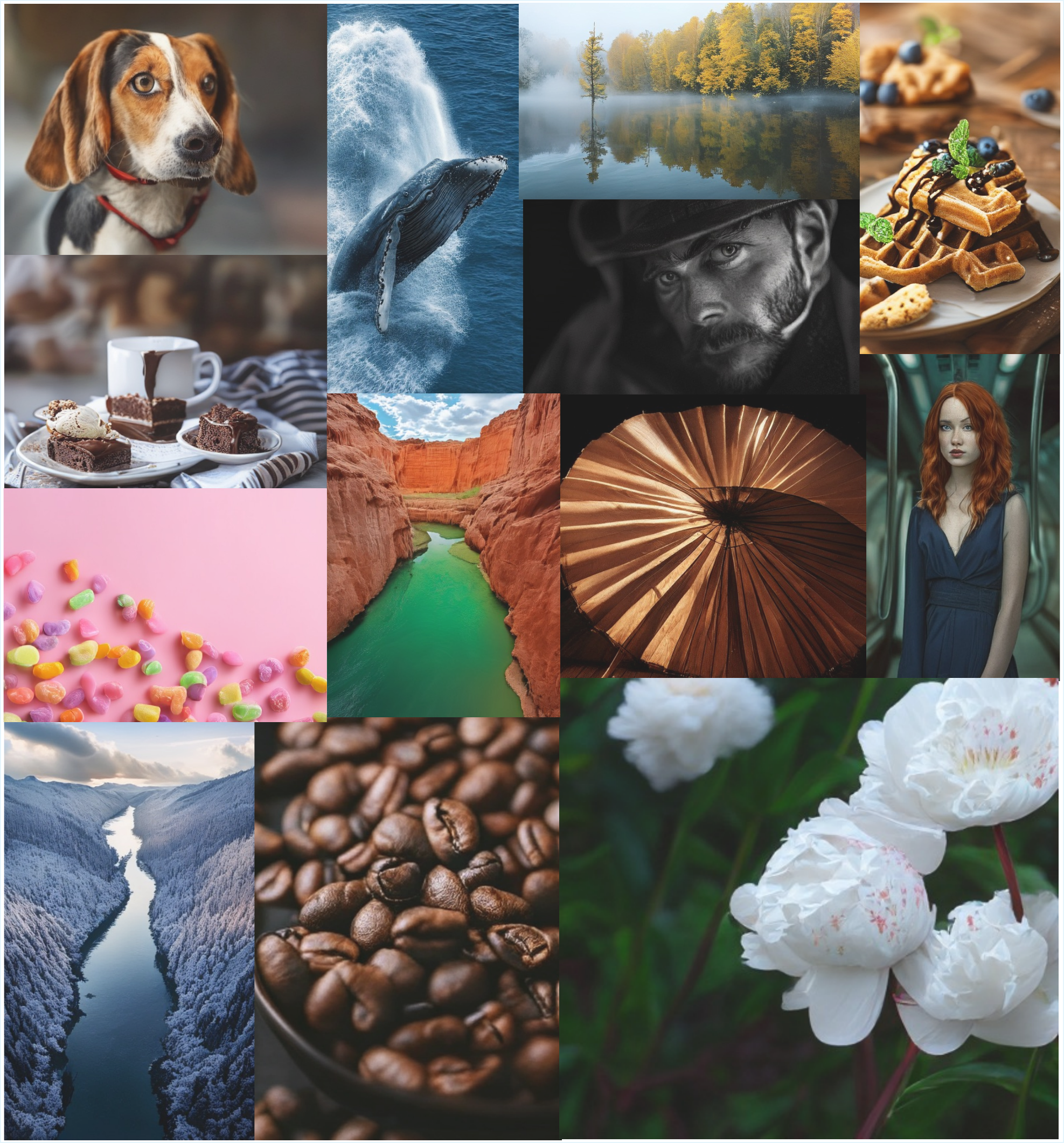}}
	\caption{Some text-to-image generation showcases of our Wallaroo.}
	\label{fig:generation_case}
    \vskip -0.25in
\end{figure}

Therefore, in this work, we adopt a vanilla next-token prediction paradigm and propose a simple autoregressive baseline called \textbf{Wallaroo}, which unifies multi-modal understanding, image generation, and editing simultaneously. Specifically, Wallaroo is built on Qwen2.5 VL~\citep{bai2025qwen2} and follows Janus~\citep{wu2025janus} to decouple the visual encoding into different pathways for understanding and generation, respectively. We employ an elaborate four-stage strategy to preserve its exceptional multi-modal understanding performance, all while endowing the model with generation and editing capability. Moreover, attributing to our subtle multi-resolution training tricks and bilingual training dataset, Wallaroo supports multi-resolution image input and output as well as bilingual language for both Chinese and English as shown in Fig~\ref{fig:generation_case}.

We conduct extensive experiments on various benchmarks to evaluate Wallaroo's capability. The results show that our model yields competitive performance and even exceeds other counterparts, implying the potential of autoregressive in unifying multi-modality understanding and generation. Our contribution can be summarized as follows:

$\bullet$ To the best of our knowledge, Wallaroo is one of the pioneering efforts that  leverage next-token prediction to unify multi-modal understanding, image generation, and editing within a simple autoregressive model.

$\bullet$ Wallaroo supports multi-resolution image input and output as well as bilingual language for both Chinese and English.

$\bullet$ Extensive experimental results show that Wallaroo produces competitive performance and even exceeds other counterparts,  implying the promising potential of autoregressive in unifying multi-modality understanding and generation.

\section{Related Work}

Unifying multi-modal understanding and generation shows attractive vision on the way to artificial general intelligence. This field has recently seen the emergence of numerous intriguing work. We roughly categorize them into three classes.

\textbf{Multi-modal Understanding Models as Enhanced Conditional Encoders.} These efforts~\citep{pan2025transfer, zeng2026forge, wu2025omnigen2, chen2025blip3, wang2025ovis, kou2024orthus, lin2025uniworld} replace traditional text encoders like T5 and CLIP with multi-modal understanding models due to their superior capabilities. For example, MetaQueries~\citep{pan2025transfer} connects the latents from multi-modal models to the diffusion decoder through learnable queries, enabling knowledge-augmented image generation. Blip3-o~\citep{chen2025blip3} further leverages diffusion models to regressive conditional representations for following image generation. Different from Blip3-o, Orthus~\citep{kou2024orthus} uses a single multi-modal model to jointly encode text and image conditions and employs a patch-level diffusion for generation following MAR~\citep{li2024autoregressive}. Similar to our Wallaroo, OmniGen2~\citep{wu2025omnigen2} also enables multi-modal understanding, image generation, and editing, but it still uses multi-modal models for condition encoding. Though these efforts currently show superior performance in both understanding and generation, essentially it is a variant of a diffusion generation model. 
The unidirectional flow of information, from understanding to generation, inevitably restricts further progress.

\textbf{Integrating Autoregressive and Diffusion within Transformers.} To break unidirectional flow and leverage the advantage of both autoregressive and diffusion, some efforts integrate them into transformers in parallel. Transfusion~\citep{zhou2024transfusion} combines the next-token prediction with diffusion to train a single transformer over mixed-modality sequences. JanusFlow~\citep{ma2025janusflow} leverages rectified flow for generation within the large language model and decouples the understanding and generation encoders. Recently, Bagel~\citep{deng2025emerging} employs two transformers for understanding and generation, respectively, while facilitating information sharing through attention modules. Overall, these methods effectively enable information interaction between multi-modal understanding and generation and seem feasible from the view of performance. However, this manner requires careful design of the attention mask and provides relatively low information interaction efficiency due to the existence of noise representation in diffusion.

\textbf{Unifying Understanding and Generation via Next-token Prediction.} Autoregressive models offer an alternative to breaking the unidirectional flow, while subtly preventing noise representations from reducing interaction efficiency. Chameleon~\citep{team2024chameleon} is a key prior effort that fully leverages autoregressive models to unify multi-modal understanding and generation. However, the poor performance of visual tokenizer  restricts the model's performance. To alleviate it, TokenFlow~\citep{qu2025tokenflow} and UniTok~\citep{ma2025unitok} enhance the performance of visual tokenizer by bridging the representation gap between multi-modal understanding and generation. Differing from these methods, Janus~\citep{wu2025janus} decouples visual encoding into separate pathways and alleviates the conflict between the visual encoder’s roles in understanding and generation. Janus-Pro~\citep{chen2025janus} further scales the model and data to obtain better performance. OneCAT~\cite{li2025onecat} also uses an autoregressive model while employing multiple experts for different modality and next-scale prediction for generation. Different from OneCAT, our Wallaroo employs a pure transformer with next-token prediction to unify multi-modal understanding, generation, and editing simultaneously. Hence, Wallaroo could be regarded as a vanilla baseline.


\section{Wallaroo}

\begin{figure}[t]
\centering{\includegraphics[width=1\linewidth]{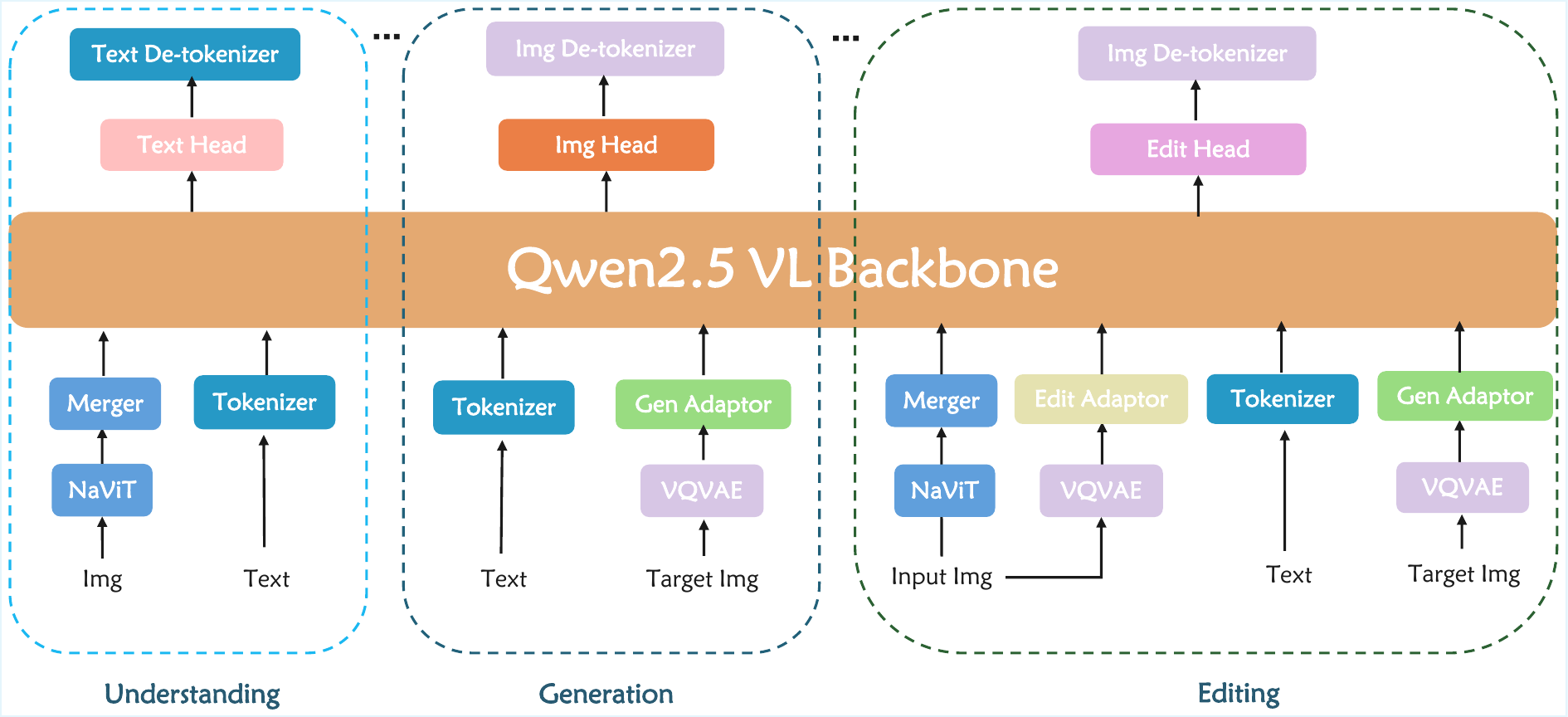}}
	\caption{Illustration of our Wallaroo. We decouple visual encoding into separate pathways for visual understanding and image generation. For editing, we integrate two complementary types of visual representations to improve Wallaroo's performance.}
	\label{fig:overview}
\end{figure}

\subsection{Architecture} 

The architecture of Wallaroo is illustrated in Fig~\ref{fig:overview}. Overall, we adopt Qwen2.5 VL as the backbone and build our Wallaroo following a minimalist principle: making as few modifications to the model as possible. Therefore, we maintain all designs in Qwen2.5 VL and use built-in NaViT to encode input images for multi-modal understanding. 

For image generation, considering task discrepancy, we additionally add a VQ tokenizer from LlamaGen~\citep{sun2024autoregressive} to convert images into discrete IDs and flatten them into 1-D. In this way, visual encoding is decoupled into different pathways. Then, we employ a generation MLP adaptor to project the codebook embeddings corresponding to each ID to align with the transformer dimension. These projected representations along with text embeddings are subsequently fed into transformer blocks for processing. Similar to multi-modal understanding, we also leverage a generation head for image discrete ID predictions. 

Interestingly, though we highlight the importance of decoupling visual encoding, for image editing, we collectively employ built-in NaViT in Qwen2.5 VL and the VQ tokenizer to encode input image to provide both semantic and low-level representations. Note that we fail to see this task in previous unified autoregressive next-token-prediction models. From the perspective of input representation, we speculate that image editing appears to be an effective link that bridges understanding and generation, which may be worth more exploration in the future (We give a detailed discussion in Sec~\ref{sec:discussion}). For VQ encoding, we use the representations from VQ \textit{encoder} instead of the quantization layer to preserve more low-level details. Considering the potential discrepancies in representations, we introduce an editing MLP adaptor to align the representations, rather than reusing the generation adaptor. For discrete ID predictions, we introduce a new edit head as we find that reusing the generation head leads to loss conflict during training. 

%


\begin{figure}[h]
\centering{\includegraphics[width=1\linewidth]{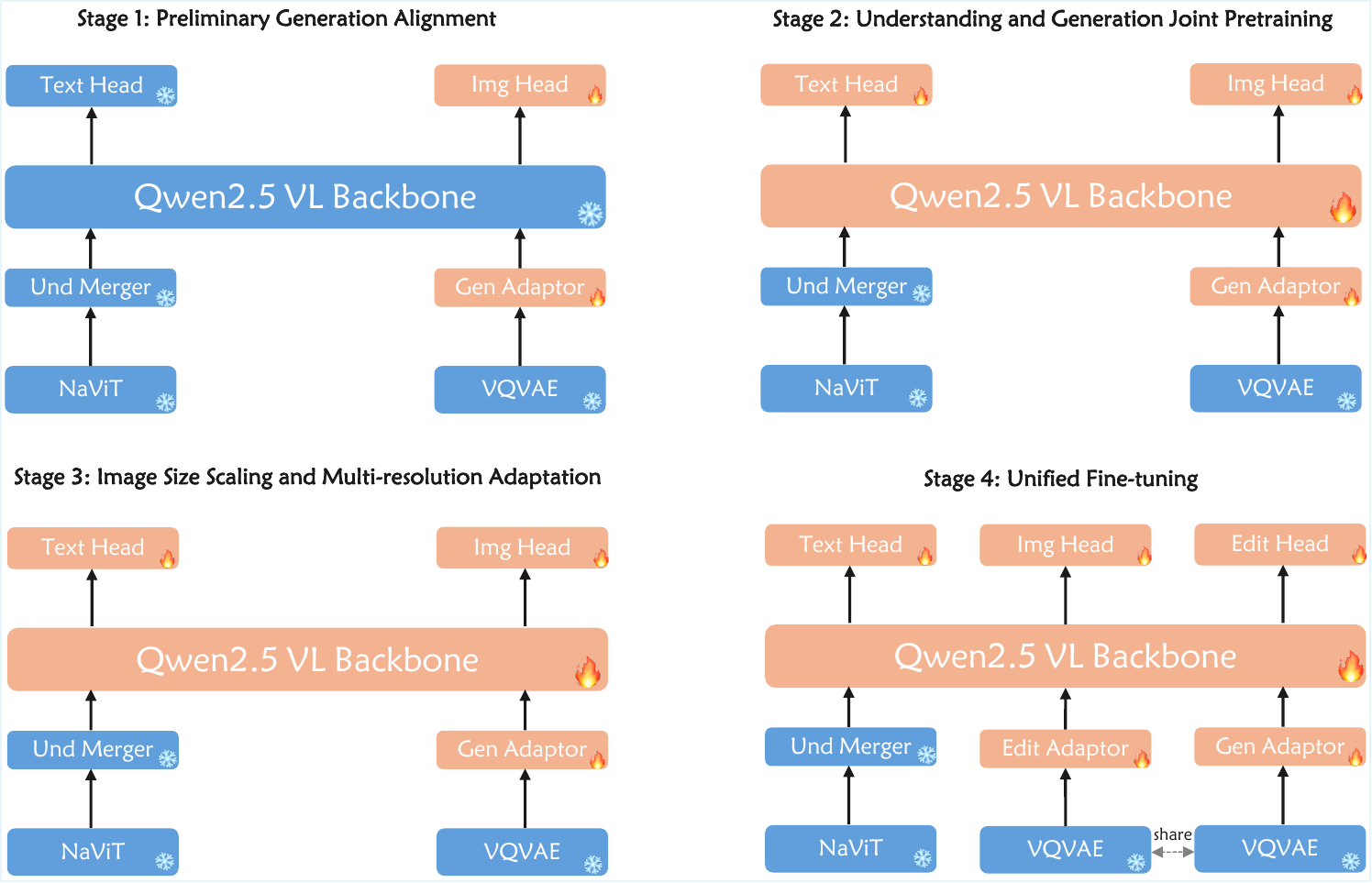}}
	\caption{A four-stage training procedure of our Wallaroo based on Qwen2.5 VL. We use flame symbols to denote modules that update their parameters, and snowflake symbols to denote modules that keep their parameters fixed.}
	\label{fig:training strategy}
\end{figure}

\subsection{Training Procedure}

To unify multi-modal understanding, generation, and editing, we design a four-stage training strategy for Wallaroo as shown in Fig~\ref{fig:training strategy}. We also give a detailed description below.

\textbf{Stage 1: Preliminary Generation Alignment.} We train the newly added generation MLP adaptor and generation head while freezing rest parameters to preliminarily align them with Qwen2.5 VL representation space. This stage also aims to endow the model with simple generation capability.

\textbf{Stage 2: Understanding and Generation Joint Pretraining.} In this stage, by utilizing multi-modal understanding data and text-to-image data, we perform a joint pretraining to further align representation space. On the one hand, we attempt to maintain Qwen2.5 VL multi-modal understanding ability. On the other hand, we enhance its generation capability. We unfreeze and fine-tune the whole model except NaViT and VQ tokenizer.

\textbf{Stage 3: Image Size Scaling and Multi-resolution Adaptation.} We first increase the image size from 384$\times$384 to 512$\times$512 and continue training for around 50K to help model better adopt following multi-resolution training. After that, we start our multi-resolution training with multi-resolution images centered around 512$\times$512. Specifically, we append two special tokens "<hw\_info>" to the end of text prompt to tell Wallaroo the needed height and weight of generated image~\footnote{We also consider other strategies, \eg, adding two special tokens indicating row and column or directly using text such as 'generate an image with a height of 256 and a width of 512'. Our experiments show that using ``<hw\_info>" slightly outperforms the other two alternatives}. Additionally, to assist our model in learning multi-resolution generation, we explicitly append an "<eol>" token (end of line) at the end of each row of image tokens to signify the line break.

\textbf{Stage 4: Unified Fine-tuning.} In this stage, we use elaborate fine-tuning dataset to further enhance its overall capability. Meanwhile, thanks to extensive pretraining that greatly enhances the model's generation capability, we use a small set of high-quality editing datasets to activate its editing functionality~\citep{wang2025image}. In other words, we fine-tune Wallaroo jointly on three tasks including multi-modal understanding, image generation, and editing.

\subsection{Training Objective}

We simply adopt next-token prediction loss as follows to optimize our model:

\begin{equation}
    L = -\sum_{i=1} \log P_{\theta}(x_{i}|x_{<i})
\end{equation}

$P(|)$ is the conditional probability of our model parameterized by $\theta$. To balance model capability, we assign the same loss weights, \ie, $1$, to all three tasks.

\subsection{Inference}

During inference, we use next-token prediction manner and adopt different heads for corresponding tasks. Specifically, we use built-in text head of Qwen2.5 VL for multi-modal understanding. We use newly added generation head for image generation and editing. Similar to Janus-Pro, we leverage classifier-free guidance (CFG) to improve generation quality: for each token,  $l_{c} = l_{u} + \gamma \cdot (l_{c}-l_{u})$, where $l_{c}$ is the conditional logit of, $l_{u}$ is the unconditional logit, and $\gamma$ is the scale for the classifier-free guidance. In this work, we set $\gamma = 3$ if not specified.

\section{Experiments}

\setlength{\tabcolsep}{0.3cm}{\begin{table}[h]
\centering
\caption{Detailed hyperparameters of each training stage. Data ratio refers to the ratio of multimodal
understanding data, visual generation data, and editing data in a batch size.}
\begin{tabular}{l | ccc c}
\toprule
\textbf{Hyperparameters} & \textbf{Stage 1} & \textbf{Stage 2} & \textbf{Stage 3} & \textbf{Stage 4} \\
\midrule
Learning rate & \(1 \times 10^{-4}\) & \(1 \times 10^{-4}\) & \(4 \times 10^{-5}\)/\(1 \times 10^{-5}\) & \(1 \times 10^{-5}\) \\
LR scheduler & Constant & Constant & Constant  & Constant \\
Weight decay & 0.0 & 0.0 & 0.0 &  0.0 \\
Gradient clip & 1.0 & 1.0 & 1.0  & 1.0  \\
Optimizer & \multicolumn{4}{c}{AdamW (\(\beta_1 = 0.9, \beta_2 = 0.95\), \(\epsilon=1e-8\))} \\
Warm-up steps & 1000 & 5000 & 0 & 0 \\
Training steps & 30K & 300K & 50K/50K & 55K \\
Batch size & 256 & 320 & 384 & 384 \\
Data Ratio & \(0:1:0\) & \(1:0:4\) & \(1:0:2\) &  \(1:1:1\) \\
\bottomrule
\end{tabular}
\label{tab:training details}
\end{table}

\subsection{Implementation Details}

We adopt Qwen2.5 VL 7B Instruct as our backbone and set the max sequence length to 4096. The VQ tokenizer has a codebook of size 16, 384 and downsamples images by a factor of 16. All generation adaptor, editing adaptor, generation head, and edit head are two-layer MLPs. We provide the detailed hyperparameter settings of each stage in Tab~\ref{tab:training details}. For multi-modal understanding data, we follow the image processing of Qwen2.5 VL. For visual generation data, in stage 1 and stage 2, we resize the short side to 384 and apply a center crop. In stage 3 and stage 4, for generation and editing, we resize a given image to the most suitable ratio from our ratio settings. For editing, we randomly mask 60\% tokens from VQ encoding to prevent the model from simply copying and pasting following~\cite{chen2025sharegpt}. In a batch, we leverage all types of data according to our data ratio. Our Wallaroo is trained on 8 nodes with each containing 8 H800 GPUs.

\subsection{Training Data}

Below, we provide the information of training data we used in each stage.

\textbf{Stage 1.} Following Pixart~\citep{chen2023pixart} and Janus-Pro~\citep{chen2025janus}, we use ImageNet1K~\citep{imagenet15russakovsky} for preliminary visual generation. We leverage ChatGPT to create multiple English/Chinese prompt templates and randomly choose one to pair with an ImageNet1K category name, \eg, "Generate an image based on the prompt: <category name>".

\textbf{Stage 2.} In this stage, we perform a joint multi-modal understanding and image generation pretraining. For understanding, we use the multi-modal datasets including LLaVA-NeXT-Data, LLAVA-OneVision-Data, M4-Instruct-Data,  QA video data (less than 60s) from LLaVA-Video-178K, \etc. We show the detailed information about the understanding datasets we used in Tab~\ref{tab:understanding-data}. Therefore, there are totally around 12M data samples. For image generation, we use in-house data.

\textbf{Stage 3.} We continue our joint multi-modal understanding and image generation pretraining in this stage. For understanding, we leverage the 12M MAmmoTH-VL dataset~\cite{guo2025mammoth}. For image generation, we use in-house data.

\setlength{\tabcolsep}{0.15cm}{\begin{table}[t]
\centering
\caption{Detailed information about the understanding datasets we used in Stage 2 and Stage 3.}
\footnotesize
\begin{tabular}{l | ccc }
\toprule
Dataset & Type & Num & Source  \\
\midrule
LLAVA-OneVision & Modality & 3M & \url{lmms-lab/LLaVA-OneVision-Data} \\
Llama-Nemotron-VLM & Modality & 2.1M & \url{nvidia/Llama-Nemotron-VLM-Dataset-v1} \\
GQA (Balanced) & Modality & 943K & \url{lmms-lab/GQA} \\
MMPR-v1.2 &  Modality & 815K & \url{OpenGVLab/MMPR-v1.2} \\
LLAVA-Next & Modality & 779K & \url{lmms-lab/LLaVA-NeXT-Data} \\
llava-v1\_5-mix & Modality & 665k & \url{liuhaotian/LLaVA-Instruct-150K} \\
M4-Instruct &  Modality & 616K & \url{lmms-lab/M4-Instruct-Data} \\
LLaVA-CC3M-Pretrain & Modality & 594K & \url{liuhaotian/LLaVA-CC3M-Pretrain-595K}  \\
llava-en-zh & Modality & 315K & \url{BUAADreamer/llava-en-zh-300k} \\
videochat2 & Modality & 233K & \url{OpenGVLab/VideoChat2-IT} \\
LLaVA-Video &  Modality &  190K & \url{lmms-lab/LLaVA-Video-178K}  \\
food-visual-instructions & Modality & 131K & \url{AdaptLLM/food-visual-instructions} \\
llava-critic & Modality & 113K & \url{lmms-lab/llava-critic-113k} \\
IconQA & Modality & 107K & \url{https://iconqa.github.io/index.html} \\
RICO-ScreenQA & Modality & 86K & \url{rootsautomation/RICO-ScreenQA} \\
clevr\_count & Modality & 70K & \url{BUAADreamer/clevr_count_70k} \\
multimodal-vqa & Modality & 76K & \url{GenAIDevTOProd/multimodal-vqa-self-instruct-enriched} \\
llava-med-zh-instruct & Modality & 57K & \url{BUAADreamer/llava-med-zh-instruct-60k} \\ 
Infinity-Instruct &  Text & 1.4M & \url{BAAI/Infinity-Instruct/7M_core} \\
commonsense\_qa & Text & 12K & \url{tau/commonsense_qa} \\
\bottomrule
\end{tabular}
\label{tab:understanding-data}
\end{table}

\textbf{Stage 4.} In this stage, for understanding, we use in-house multi-modal understanding data and part of data from LLaVA-OneVision-1.5 Instruction~\cite{an2025llava}. For generation, we use the instruction-tuning BLIP3o-60k~\citep{chen2025blip3} data from Blip3-o, text-to-image data from ShareGPT-4o-Image~\citep{chen2025sharegpt}, and text-to-image data from OpenGPT-4o-Image~\cite{chen2025opengpt}. For editing, we leverage the in-house editing data, image-to-image data from ShareGPT-4o-Image, OpenGPT-4o-Image, and GPT-Image-Edit-1.5M~\citep{wang2025gptimageedit15mmillionscalegptgeneratedimage}.

\subsection{Results}

\subsubsection{Results on Multi-modal Understanding}

To evaluate the performance of our model, we conduct extensive experiments and compare with other state-of-the-art methods in Tab~\ref{tab:multi-modal} on various multi-modal benchmarks including POPE~\citep{li2023evaluating}, MME~\citep{zhang2021mme},  MMB~\citep{liu2024mmbench}, SEED~\citep{li2023seed}, GQA~\citep{hudson2019gqa}, MMMU~\citep{yue2024mmmu}, and MM-Vet~\citep{yu2023mm}.

It can be seen that our Wallaroo obtains competitive performance compared to Qwen2.5 VL and outperforms most of previous state-of-the-art methods. For example, Wallaroo produces 83.0 for MMB, outperforming Janus-Pro, Mogao, OmniGen2, \etc. These results demonstrate the potential of autoregressive next-token prediction. On the other hand, these results also show that integrating generation into a multi-modality understanding model may lead to  a certain degree of performance degradation, suggesting that there is a long way to go to achieve mutual benefit for both tasks.

\setlength{\tabcolsep}{0.1cm}{\begin{table}[t]
\centering
\caption{Comparison with state-of-the-art unified model on multi-modal understanding benchmarks. $^{*}$ indicates that we use the \href{https://github.com/open-compass/VLMEvalKit}{VLMEvalKit} to evaluate the results.}
\begin{tabular}{lccccccccc}
\toprule
Model & Params & POPE$\uparrow$ & MME-P$\uparrow$ & MMB$\uparrow$ & SEED$\uparrow$ & GQA$\uparrow$ & MMMU$\uparrow$ & MM-Vet$\uparrow$ \\
\midrule
\multicolumn{6}{l}{\footnotesize \emph{Multi-modal Understanding Models as Enhanced Conditional Encoders}:}\\
MetaQuery & 7B + 1.6B   & - & 1685.2 & 83.5 & 76.9  &  & 58.6 & 66.6 \\
Blip3-o & 7B + 1.4B  & - & 1682.6 & 83.5 & 77.5 &  & 50.6 & 66.6\\
Ming-Lite-Uni &  8B+1.6B  & - & - & 80.7  & - & - & 51.2 &  72.3 \\
UniWorld-V1 & 7B + 12B & - & - & 83.5 & - & - & 58.6 &  67.1 \\
OmniGen2 & 3B + 4B & - & - & 79.1 & - & - & 53.1 &  61.8 \\
\midrule
\multicolumn{6}{l}{\footnotesize \emph{Integrating Autoregressive and Diffusion within Transformers}:}\\
Show-o & 1.3B & - & 1097.2 & - & 51.5 &  58.0 & 27.4 & -\\
JanusFlow & 1.3B &  88.0  & 1333.1  & 74.9 & 70.5  & 60.3  & 29.3 & 30.9   \\
Show-o2 &  7B & - & 1620.5 & 79.3 & 69.8 & 63.1 & 48.9 & - \\
Mogao & 7B & -& 1592.0 & 75.0 & 74.6 & 60.9 & 44.2 & -  \\
BAGEL & 7B+7B & - & 1687 &  85.0 & - & - & 55.3 & 67.2   \\
\midrule
\multicolumn{6}{l}{\footnotesize \emph{Unifying Understanding and Generation via Next-token Prediction}:}\\
Chameleon & 7B & - & - & - & - & - & 22.4 & 8.3 \\
Emu3 & 8B & - & - & 58.5  & 68.2 & 60.3 & 31.6 &  37.2 \\
TokenFlow &  13B & 86.8 & 1545.9 & 68.9 & 68.7 & 62.7 & 38.7 & 40.7 \\
VILA-U & 7B & 85.8 & 1401.8 & - & 59.0 & 60.8 & - & 33.5 \\
Janus &  1.5B & 87.0 &  1338.0 & 69.4 & 63.7 &  59.1 & 30.5 & 34.3 \\
Janus-Pro  & 7B & 87.4 & 1567.1 & 79.2 &  72.1 & 62.0 & 41.0 & 50.0 \\
\midrule

Qwen2.5 VL$^*$ & 7B  & 86.3 & 1692.5 & 83.0 & 77.1 & 60.3 & 44.9 & 62.1 \\
\textbf{Wallaroo}$^*$ & 7B  & 86.4  & 1690.3 & 83.0 & 76.4 & 60.1 & 42.7 & 50.1 \\ 
\bottomrule
\end{tabular}
\label{tab:multi-modal}
\end{table}

\subsubsection{Results on Image Generation}

\textbf{Results on GenEval.} We evaluate the text-to-image generation performance of our model on GenEval benchmark~\citep{ghosh2023geneval}. As shown in Tab~\ref{tab:geneval}, we can see that Wallaroo could produce competitive results compared to Janus-Pro and Show-o2, suggesting the promising potential of pure autoregressive next-token prediction in image generation even when unifying three tasks within a single model. At the same time, we must acknowledge that Wallaroo falls short compared to diffusion-based models like OmniGen2 and BAGEL. This result is reasonable as vector quantization in VQ encoding leads to significant loss of image details, whereas diffusion models do not experience this issue.

\setlength{\tabcolsep}{0.1cm}{\begin{table}[h]
\centering
\caption{Comparison of text-to-image generation ability on GenEval benchmark.}
  \begin{tabular}{lccccccccc}
    \toprule
     Method & Single Obj. & Two Obj. & Counting & Colors & Position & Color Attri. & Overall$\uparrow$ \\
    \midrule
    \multicolumn{6}{l}{\footnotesize \emph{Multi-modal Understanding Models as Enhanced Conditional Encoders}:}\\
     MetaQuery$^{*}$ & - & - & - & - & - & - & 0.80  \\
     Blip3-o$^{*}$ & - & - & - & - & - & - & 0.84 \\
     Ming-Lite-Uni & 0.99 & 0.76 & 0.53 & 0.87 & 0.26 & 0.30 & 0.62 \\
     UniWorld-V1 & 0.99 & 0.93 & 0.79 & 0.89 & 0.49 & 0.70 & 0.80 \\
     OmniGen2 & 1 & 0.95 & 0.64 & 0.88 & 0.55 & 0.76 & 0.80 \\
     \midrule
     \multicolumn{6}{l}{\footnotesize \emph{Integrating Autoregressive and Diffusion within Transformers}:}\\
     Show-o & 0.95 & 0.52 & 0.49 & 0.82 & 0.11 & 0.28 & 0.53 \\
     JanusFlow  & 0.97 & 0.59 & 0.45 & 0.83 & 0.53 & 0.42 & 0.63 \\
     Show-o2 & 1.00 & 0.87 & 0.58 &  0.92 & 0.52 & 0.62 & 0.76 \\
     Mogao$^{*}$ & 1.00 & 0.97 & 0.83 & 0.93 & 0.84 & 0.80 & 0.89 \\
     BAGEL & 0.99 & 0.94 & 0.81 & 0.88 & 0.64 & 0.63 & 0.82 \\
     \midrule
     \multicolumn{6}{l}{\footnotesize \emph{Unifying Understanding and Generation via Next-token Prediction}:}\\
     Chameleon & - & - & - & - & - & - & 0.39 \\
     Emu3  & - & - & - & - & - & - &  0.66 \\
     TokenFlow & 0.95 & 0.60 & 0.41 & 0.81 & 0.16 & 0.24 & 0.55 \\
     Janus &  0.97 & 0.68 & 0.30 & 0.84 & 0.46 & 0.42 & 0.61 \\
     Janus-Pro & 1.00 & 0.85 & 0.53 & 0.90 & 0.69 & 0.58 & 0.76\\
     Janus-4o & 1.00 & 0.92 & 0.58 & 0.88 & 0.70 & 0.70 & 0.80 \\
     \midrule
     \textbf{Wallaroo} & 1.00 & 0.81 & 0.51 & 0.87 & 0.69 & 0.61 & 0.75 \\
    \bottomrule
  \end{tabular}
\label{tab:geneval}
\end{table}

\textbf{Results on DPG.} To further show the text-to-image capability of our Wallaroo, we conduct experiments on DPG benchmark~\citep{hu2024ella} and report the results in Tab~\ref{tab:dpg}. Similar to the observation in Geneval benchmark, our Wallaroo yields competitive result compared to JanusFlow and EMU3. However, one may also notice that Wallaroo is inferior to Janus-Pro and Janus-4o. This result may potentially be due to the data bias involved in our training.

\setlength{\tabcolsep}{0.2cm}{\begin{table}[h]
\centering
\caption{Comparison of text-to-image generation ability on DPG benchmark. We use CFG=2.5.}
  \begin{tabular}{llcccccccc}
    \toprule
     Method & Global & Entity & Attribute & Relation & Other & Overall$\uparrow$ \\
    \midrule
    \multicolumn{6}{l}{\footnotesize \emph{Multi-modal Understanding Models as Enhanced Conditional Encoders}:}\\
     MetaQuery & - & - & - & - & - & 82.05 \\
     Blip3-o & - & - & - & - & - & 81.60 \\
    UniWorld-V1 & 83.64 & 88.39 & 88.44 & 89.27 & 87.22 & 81.38 \\
    OmniGen2 & 88.81 & 88.83 & 90.18 & 89.37 & 90.27 & 83.57 \\
    \midrule
    \multicolumn{6}{l}{\footnotesize \emph{Integrating Autoregressive and Diffusion within Transformers}:}\\
    Show-o & 79.33 & 75.44 & 78.02 & 84.45 & 60.80 & 67.27 \\
    JanusFlow & 87.03 & 87.31 & 87.39 & 89.79 & 88.10 & 80.09 \\
    Show-o2 & 89.00 & 91.78 & 89.96 & 91.81 & 91.64 & 86.14 \\
    Mogao & 82.37 & 90.03 & 88.26 & 93.18 & 85.40 & 84.33 \\
    BAGEL & 88.94 & 90.37 & 91.29 & 90.82 & 88.67 & 85.07 \\
    \midrule
    \multicolumn{6}{l}{\footnotesize \emph{Unifying Understanding and Generation via Next-token Prediction}:}\\
    EMU3 & 85.21  &86.68 & 86.84 & 90.22 & 83.15 & 80.60 \\
    TokenFlow & 78.72 & 79.22 & 81.29 & 85.22 & 71.20 & 73.38 \\
    Janus & 82.33 & 87.38 & 87.70 & 85.46 & 86.41 & 79.68 \\
    Janus-Pro & 86.90 & 88.90 & 89.40 & 89.32 & 89.48 & 84.19 \\
    Janus-4o & 92.59 & 90.61 & 89.51 & 91.77 & 89.01 & 85.71 \\
    \midrule
    \textbf{Wallaroo} & 75.00 & 81.20 & 83.33 & 78.13 & 92.31 & 79.35 \\
    \bottomrule
  \end{tabular}
\label{tab:dpg}
\end{table}

\subsubsection{Results on Image Editing}

\textbf{Results on ImgEdit.} We also evaluate the editing performance of our model on ImgEdit benchmark~\citep{ye2025imgedit}. As shown in Tab~\ref{tab:imageedit}, Wallaroo obtains 2.92 overall performance, catching even outperforming most pure image generation/editing models including AnyEdit, UltraEdit, and OmniGen. Additionally, we find that the editing performance of Wallaroo is inferior to that of diffusion-based unified models such as BAGEL, UniWorld-V1, and OmniGen2. Similar to image generation, the results are due to the limitation of generation paradigm. One may also notice that Janus-4o, which adopts the same autoregressive paradigm to our method, outperforms Wallaroo. We speculate that this is because Janus-4o forgoes multimodal understanding and concentrates exclusively on generation and editing while our Wallaroo highlights the equal importance of understanding, generation and editing.

\setlength{\tabcolsep}{0.1cm}{\begin{table}[h]
\centering
\caption{Comparison of image editing capability on ImgEdit benchmark.}
\small
\begin{tabular}{l | c c c c c c c c c c}
\toprule
Method & Extract & Adjust & Background & Add & Replace & Remove & Style & Compose & Action & Overall$\uparrow$ \\
\midrule
 AnyEdit & 1.88 & 2.95 & 2.24 &  3.18 & 2.47 & 2.23 &  2.85 &  1.56 & 2.65 & 2.45 \\
 UltraEdit &  2.13 & 2.81 & 2.83 & 3.44 & 2.96 & 1.45 & 3.76 & 1.91 & 2.98 & 2.70 \\
 OmniGen &  1.71 & 3.04 & 3.21 & 3.47 & 2.94 & 2.43 &  4.19 & 2.24 & 3.38 & 2.96  \\
 Step1X-Edit & 1.76 & 3.14 & 3.16 & 3.88 &  3.40 &  2.41 &  4.63 & 2.64 & 2.52 & 3.06 \\
 BAGEL & 1.70 & 3.31 &  3.24 &  3.56 & 3.30 & 2.62 & 4.49 &  2.38 & 4.17 & 3.20 \\
 UniWorld-V1 &  2.27 & 3.64 & 2.99 & 3.82 & 3.47 & 3.24 & 4.21 & 2.96 & 2.74 & 3.26 \\
  Janus-4o & 2.28 & 4.13 & 3.32 & 3.60 & 3.27 & 2.28 & 4.47 & 4.47  &  2.74 & 3.26 \\
 OmniGen2 & 1.77 & 3.06  & 3.57 & 3.57 & 3.74 & 3.20  & 4.81 & 2.52 & 4.68 & 3.44 \\
     \midrule
    \textbf{Wallaroo} & 2.02 & 3.41 & 2.93 & 3.32 & 2.54 & 1.61 & 4.14 & 2.58 & 3.73 & 2.92 \\
\bottomrule
\end{tabular}
\label{tab:imageedit}
\end{table}

\subsection{Ablation Studies}

\textbf{VQ Tokenizer Selection.} Besides LlamaGen, we also consider the tokenizer from MoVQGAN~\citep{zheng2022movq}, which is a 8$\times$8 downsampling VQ tokenizer while keeping the same codebook size to LlamaGen. In Tab~\ref{tab:vq}, we compare the generation performance of different VQ tokenizer on ImageNet1K in stage 1 under the same training steps using a 3B Qwen2.5 VL model. The results show that LlamaGen is superior over MoVQGAN in all metrics. We hypothesize that MoVQGAN generates more tokens because of its smaller downsampling setting, which in turn leads to slower convergence compared to LlamaGen. Considering that we will scale image size in following stage, to save training time, we use LlamaGen as our default VQ tokenizer.  

\setlength{\tabcolsep}{0.5cm}{\begin{table}[h]
\centering
\caption{Comparison of different VQ tokenizer on generation performance.}
\begin{tabular}{l | ccc c}
\toprule
VQ Tokenizer & Inception Score & FID & sFID \\
\midrule
LlamaGen & 199.06 & 11.04 & 15.04  \\
MoVQGAN & 136.10 & 14.72 & 25.36  \\
\bottomrule
\end{tabular}
\label{tab:vq}
\end{table}

\textbf{Different Mask Ratios for Editing.}
To prevent the model from simply copying and pasting, we randomly mask a fixed ratio of content during training. As shown in Tab~\ref{tab:mask}, we evaluate the influence of different mask ratios on editing performance on ImageEdit benchmark using a 3B Qwen2.5 VL model. One can see that when mask ratio is set to 0.6, the model achieves the best performance among all mask ratio settings. We consider 0.6 to be an effective trade-off ratio, balancing model regularization with the provision of sufficient low-level representations.

\setlength{\tabcolsep}{0.08cm}{\begin{table}[h]
\centering
\caption{Comparison of different mask ratio on editing performance.}
\begin{tabular}{l | c c c c c c c c c c}
\toprule
Ratio & Extract & Adjust & Background & Add & Replace & Remove & Style & Compose & Action & Overall \\
\midrule
0.5 & 1.86 & 2.63 & 2.93 & 2.92 & 2.39 & 1.33 & 4.06 & 1.91 & 2.77 & 2.53  \\
0.6 & 2.05 & 3.19 & 2.88 & 2.74 & 2.22 & 1.41 & 4.44 & 2.09 & 3.08 & 2.67 \\
0.75 & 1.82 & 2.61 & 2.98 & 2.94 & 2.26 & 1.41 & 3.91 & 1.83 & 2.76 & 2.50 \\
\bottomrule
\end{tabular}
\label{tab:mask}
\end{table}

\section{Discussion} \label{sec:discussion}

As we have mentioned above, employing an autoregressive model to unify understanding and generation is an effective method that enables lossless representation interaction compared to other two paradigms. However, a persistent issue is that vector quantization in VQ encoding causes substantial loss of image details, thereby constraining the quality of image generation. There could be two ways to alleviate this issue. We could leverage diffusion models as a post-processing step to refine the output image (in pixel/latent space). Another way is to train a more powerful VQ tokenizer, \eg, scaling tokenizer size and designing better quantization methods.

So far, it remains unclear whether multi-modal understanding and generation mutually enhance each other in autoregressive models. The primary issue is the incompatibility between high-level representations from multi-modal understanding and low-level representations from generation. We hypothesize that the two types of representations rely on an intermediate medium for better interaction. Thus a natural idea is to introduce intermediate representations that bridge the gap between the two types of representations above. Another way, we speculate, is starting from \textit{editing task}. Considering that previous efforts (also including our Wallaroo) primarily start from a language model or a multi-modal understanding model, they thereby focus on how to incorporate image generation capability. What if we start from an autoregressive editing model? This editing model takes both high-level and low-level representations as input, naturally and implicitly reconciling their conflicts.

Additionally, the type of positional encoding for different modality is critical. During our preliminary experiments, we attempt to use VQ tokenizer to encoding images to yield low-level representations for both understanding task and editing task (through the same editing adaptor to align dimension). Interestingly, we observe that adopting distinct positional encoding schemes for image representations (\eg, 1-D for editing and 2-D for understanding) enables the model to preserve image consistency and substantially enhances editing performance. However, applying the same positional encoding scheme to both tasks causes the editing task to lose image consistency, to some extent, degenerating into image generation. This contrast highlights the importance of different positional encoding.
This result also implies that low-level representations may serve as \textit{different} roles in understanding task and editing task as they need different types of positional encoding for distinguishment.

Finally, the sequence of different information is crucial for editing. In our experiment, we find that if we formulate 2-D high-level representation followed by 1-D low-level representations and 1-D text instruction, the edited image is terrible. If we reverse the sequence of image representation, \ie, 1-D low-level representation followed by 2-D high-level representation and 1-D text instruction, the editing performance is significantly improved. This result indicates that when multiple tasks are integrated into one autoregressive model, editing performance is sensitive to the sequence of token information. We leave the potential reason behind as future work.

\section{Limitation}

Wallaroo leverages three separate heads, \ie, text head, image head, and edit head, to perform multi-modality understanding and image generation/editing. As a result, users need to manually toggle the function they wish to use. To some extent, this inconvenience constrains the model's intelligence. It would be more effective if the model could dynamically choose the appropriate head based on the context. 

\section{Conclusion}

In this work, we present a simple baseline called Wallaroo. To the best of our knowledge, it is one of the pioneering efforts that unifies multi-modal understanding, generation, and editing with a pure autoregressive model through next-token prediction . It supports multi-resolution image input and output as well as bilingual language for both Chinese and English. Our extensive experiments demonstrate its competitive performance in various evaluation benchmarks, suggesting the promising potential of autoregressive in unifying multi-modality understanding and generation. Finally, we also discuss the existing issues and some findings in this research direction and propose possible solutions, hoping to inspire further efforts in the field and the creation of more extraordinary work.


\bibliography{iclr2025_conference}

@article{liu2023visual,
  title={Visual instruction tuning},
  author={Liu, Haotian and Li, Chunyuan and Wu, Qingyang and Lee, Yong Jae},
  journal={Advances in neural information processing systems},
  volume={36},
  pages={34892--34916},
  year={2023}
}

@inproceedings{rombach2022high,
  title={High-resolution image synthesis with latent diffusion models},
  author={Rombach, Robin and Blattmann, Andreas and Lorenz, Dominik and Esser, Patrick and Ommer, Bj{\"o}rn},
  booktitle={Proceedings of the IEEE/CVF conference on computer vision and pattern recognition},
  pages={10684--10695},
  year={2022}
}

@article{pan2025transfer,
  title={Transfer between modalities with metaqueries},
  author={Pan, Xichen and Shukla, Satya Narayan and Singh, Aashu and Zhao, Zhuokai and Mishra, Shlok Kumar and Wang, Jialiang and Xu, Zhiyang and Chen, Jiuhai and Li, Kunpeng and Juefei-Xu, Felix and others},
  journal={arXiv preprint arXiv:2504.06256},
  year={2025}
}

@article{chen2025blip3,
  title={Blip3-o: A family of fully open unified multimodal models-architecture, training and dataset},
  author={Chen, Jiuhai and Xu, Zhiyang and Pan, Xichen and Hu, Yushi and Qin, Can and Goldstein, Tom and Huang, Lifu and Zhou, Tianyi and Xie, Saining and Savarese, Silvio and others},
  journal={arXiv preprint arXiv:2505.09568},
  year={2025}
}

@article{wu2025omnigen2,
  title={OmniGen2: Exploration to Advanced Multimodal Generation},
  author={Wu, Chenyuan and Zheng, Pengfei and Yan, Ruiran and Xiao, Shitao and Luo, Xin and Wang, Yueze and Li, Wanli and Jiang, Xiyan and Liu, Yexin and Zhou, Junjie and others},
  journal={arXiv preprint arXiv:2506.18871},
  year={2025}
}

@article{wang2025ovis,
  title={Ovis-U1 Technical Report},
  author={Wang, Guo-Hua and Zhao, Shanshan and Zhang, Xinjie and Cao, Liangfu and Zhan, Pengxin and Duan, Lunhao and Lu, Shiyin and Fu, Minghao and Chen, Xiaohao and Zhao, Jianshan and others},
  journal={arXiv preprint arXiv:2506.23044},
  year={2025}
}

@article{lin2025uniworld,
  title={Uniworld: High-resolution semantic encoders for unified visual understanding and generation},
  author={Lin, Bin and Li, Zongjian and Cheng, Xinhua and Niu, Yuwei and Ye, Yang and He, Xianyi and Yuan, Shenghai and Yu, Wangbo and Wang, Shaodong and Ge, Yunyang and others},
  journal={arXiv preprint arXiv:2506.03147},
  year={2025}
}

@article{kou2024orthus,
  title={Orthus: Autoregressive interleaved image-text generation with modality-specific heads},
  author={Kou, Siqi and Jin, Jiachun and Liu, Zhihong and Liu, Chang and Ma, Ye and Jia, Jian and Chen, Quan and Jiang, Peng and Deng, Zhijie},
  journal={arXiv preprint arXiv:2412.00127},
  year={2024}
}

@article{li2024autoregressive,
  title={Autoregressive image generation without vector quantization},
  author={Li, Tianhong and Tian, Yonglong and Li, He and Deng, Mingyang and He, Kaiming},
  journal={Advances in Neural Information Processing Systems},
  volume={37},
  pages={56424--56445},
  year={2024}
}

@article{zhou2024transfusion,
  title={Transfusion: Predict the next token and diffuse images with one multi-modal model},
  author={Zhou, Chunting and Yu, Lili and Babu, Arun and Tirumala, Kushal and Yasunaga, Michihiro and Shamis, Leonid and Kahn, Jacob and Ma, Xuezhe and Zettlemoyer, Luke and Levy, Omer},
  journal={arXiv preprint arXiv:2408.11039},
  year={2024}
}

@inproceedings{ma2025janusflow,
  title={Janusflow: Harmonizing autoregression and rectified flow for unified multimodal understanding and generation},
  author={Ma, Yiyang and Liu, Xingchao and Chen, Xiaokang and Liu, Wen and Wu, Chengyue and Wu, Zhiyu and Pan, Zizheng and Xie, Zhenda and Zhang, Haowei and Yu, Xingkai and others},
  booktitle={Proceedings of the Computer Vision and Pattern Recognition Conference},
  pages={7739--7751},
  year={2025}
}

@article{deng2025emerging,
  title={Emerging properties in unified multimodal pretraining},
  author={Deng, Chaorui and Zhu, Deyao and Li, Kunchang and Gou, Chenhui and Li, Feng and Wang, Zeyu and Zhong, Shu and Yu, Weihao and Nie, Xiaonan and Song, Ziang and others},
  journal={arXiv preprint arXiv:2505.14683},
  year={2025}
}

@article{team2024chameleon,
  title={Chameleon: Mixed-modal early-fusion foundation models},
  author={Team, Chameleon},
  journal={arXiv preprint arXiv:2405.09818},
  year={2024}
}

@inproceedings{qu2025tokenflow,
  title={Tokenflow: Unified image tokenizer for multimodal understanding and generation},
  author={Qu, Liao and Zhang, Huichao and Liu, Yiheng and Wang, Xu and Jiang, Yi and Gao, Yiming and Ye, Hu and Du, Daniel K and Yuan, Zehuan and Wu, Xinglong},
  booktitle={Proceedings of the Computer Vision and Pattern Recognition Conference},
  pages={2545--2555},
  year={2025}
}

@article{ma2025unitok,
  title={Unitok: A unified tokenizer for visual generation and understanding},
  author={Ma, Chuofan and Jiang, Yi and Wu, Junfeng and Yang, Jihan and Yu, Xin and Yuan, Zehuan and Peng, Bingyue and Qi, Xiaojuan},
  journal={arXiv preprint arXiv:2502.20321},
  year={2025}
}

@inproceedings{wu2025janus,
  title={Janus: Decoupling visual encoding for unified multimodal understanding and generation},
  author={Wu, Chengyue and Chen, Xiaokang and Wu, Zhiyu and Ma, Yiyang and Liu, Xingchao and Pan, Zizheng and Liu, Wen and Xie, Zhenda and Yu, Xingkai and Ruan, Chong and others},
  booktitle={Proceedings of the Computer Vision and Pattern Recognition Conference},
  pages={12966--12977},
  year={2025}
}

@article{chen2025janus,
  title={Janus-pro: Unified multimodal understanding and generation with data and model scaling},
  author={Chen, Xiaokang and Wu, Zhiyu and Liu, Xingchao and Pan, Zizheng and Liu, Wen and Xie, Zhenda and Yu, Xingkai and Ruan, Chong},
  journal={arXiv preprint arXiv:2501.17811},
  year={2025}
}

@article{zhu2024mole,
  title={Mole: Enhancing human-centric text-to-image diffusion via mixture of low-rank experts},
  author={Zhu, Jie and Chen, Yixiong and Ding, Mingyu and Luo, Ping and Wang, Leye and Wang, Jingdong},
  journal={Advances in Neural Information Processing Systems},
  volume={37},
  pages={29354--29386},
  year={2024}
}

@inproceedings{chen2024internvl,
  title={Internvl: Scaling up vision foundation models and aligning for generic visual-linguistic tasks},
  author={Chen, Zhe and Wu, Jiannan and Wang, Wenhai and Su, Weijie and Chen, Guo and Xing, Sen and Zhong, Muyan and Zhang, Qinglong and Zhu, Xizhou and Lu, Lewei and others},
  booktitle={Proceedings of the IEEE/CVF conference on computer vision and pattern recognition},
  pages={24185--24198},
  year={2024}
}

@article{team2025kimi,
  title={Kimi-vl technical report},
  author={Team, Kimi and Du, Angang and Yin, Bohong and Xing, Bowei and Qu, Bowen and Wang, Bowen and Chen, Cheng and Zhang, Chenlin and Du, Chenzhuang and Wei, Chu and others},
  journal={arXiv preprint arXiv:2504.07491},
  year={2025}
}

@article{wang2024qwen2,
  title={Qwen2-vl: Enhancing vision-language model's perception of the world at any resolution},
  author={Wang, Peng and Bai, Shuai and Tan, Sinan and Wang, Shijie and Fan, Zhihao and Bai, Jinze and Chen, Keqin and Liu, Xuejing and Wang, Jialin and Ge, Wenbin and others},
  journal={arXiv preprint arXiv:2409.12191},
  year={2024}
}

@article{li2024hunyuan,
  title={Hunyuan-dit: A powerful multi-resolution diffusion transformer with fine-grained chinese understanding},
  author={Li, Zhimin and Zhang, Jianwei and Lin, Qin and Xiong, Jiangfeng and Long, Yanxin and Deng, Xinchi and Zhang, Yingfang and Liu, Xingchao and Huang, Minbin and Xiao, Zedong and others},
  journal={arXiv preprint arXiv:2405.08748},
  year={2024}
}

@article{hong2022cogvideo,
  title={Cogvideo: Large-scale pretraining for text-to-video generation via transformers},
  author={Hong, Wenyi and Ding, Ming and Zheng, Wendi and Liu, Xinghan and Tang, Jie},
  journal={arXiv preprint arXiv:2205.15868},
  year={2022}
}

@article{zeng2026forge,
  title={Forge-and-Quench: Enhancing Image Generation for Higher Fidelity in Unified Multimodal Models},
  author={Zeng, Yanbing and Wang, Jia and Ma, Hanghang and Wu, Junqiang and Zhu, Jie and Wei, Xiaoming and Hu, Jie},
  journal={arXiv preprint arXiv:2601.04706},
  year={2026}
}

@inproceedings{esser2024scaling,
  title={Scaling rectified flow transformers for high-resolution image synthesis},
  author={Esser, Patrick and Kulal, Sumith and Blattmann, Andreas and Entezari, Rahim and M{\"u}ller, Jonas and Saini, Harry and Levi, Yam and Lorenz, Dominik and Sauer, Axel and Boesel, Frederic and others},
  booktitle={Forty-first international conference on machine learning},
  year={2024}
}

@article{bai2025qwen2,
  title={Qwen2. 5-vl technical report},
  author={Bai, Shuai and Chen, Keqin and Liu, Xuejing and Wang, Jialin and Ge, Wenbin and Song, Sibo and Dang, Kai and Wang, Peng and Wang, Shijie and Tang, Jun and others},
  journal={arXiv preprint arXiv:2502.13923},
  year={2025}
}

@article{sun2024autoregressive,
  title={Autoregressive model beats diffusion: Llama for scalable image generation},
  author={Sun, Peize and Jiang, Yi and Chen, Shoufa and Zhang, Shilong and Peng, Bingyue and Luo, Ping and Yuan, Zehuan},
  journal={arXiv preprint arXiv:2406.06525},
  year={2024}
}

@article{chen2025sharegpt,
  title={ShareGPT-4o-Image: Aligning Multimodal Models with GPT-4o-Level Image Generation},
  author={Chen, Junying and Cai, Zhenyang and Chen, Pengcheng and Chen, Shunian and Ji, Ke and Wang, Xidong and Yang, Yunjin and Wang, Benyou},
  journal={arXiv preprint arXiv:2506.18095},
  year={2025}
}

@article{chen2023pixart,
  title={Pixart-alpha: Fast training of diffusion transformer for photorealistic text-to-image synthesis},
  author={Chen, Junsong and Yu, Jincheng and Ge, Chongjian and Yao, Lewei and Xie, Enze and Wu, Yue and Wang, Zhongdao and Kwok, James and Luo, Ping and Lu, Huchuan and others},
  journal={arXiv preprint arXiv:2310.00426},
  year={2023}
}

@article{imagenet15russakovsky,
    Author = {Olga Russakovsky and Jia Deng and Hao Su and Jonathan Krause and Sanjeev Satheesh and Sean Ma and Zhiheng Huang and Andrej Karpathy and Aditya Khosla and Michael Bernstein and Alexander C. Berg and Li Fei-Fei},
    Title = { {ImageNet Large Scale Visual Recognition Challenge} },
    Year = {2015},
    journal   = {International Journal of Computer Vision (IJCV)},
    doi = {10.1007/s11263-015-0816-y},
    volume={115},
    number={3},
    pages={211-252}
}

@article{li2023evaluating,
  title={Evaluating object hallucination in large vision-language models},
  author={Li, Yifan and Du, Yifan and Zhou, Kun and Wang, Jinpeng and Zhao, Wayne Xin and Wen, Ji-Rong},
  journal={arXiv preprint arXiv:2305.10355},
  year={2023}
}

@article{zhang2021mme,
  title={Mme: A comprehensive evaluation benchmark for multimodal large language models},
  author={Zhang, Yunhang Shen Yulei Qin Mengdan and Zheng, Xu Lin Jinrui Yang Xiawu and Wu, Ke Li Xing Sun Yunsheng and Fu, Rongrong Ji Chaoyou and Chen, Peixian},
  journal={arXiv preprint arXiv:2306.13394},
  year={2021}
}

@inproceedings{liu2024mmbench,
  title={Mmbench: Is your multi-modal model an all-around player?},
  author={Liu, Yuan and Duan, Haodong and Zhang, Yuanhan and Li, Bo and Zhang, Songyang and Zhao, Wangbo and Yuan, Yike and Wang, Jiaqi and He, Conghui and Liu, Ziwei and others},
  booktitle={European conference on computer vision},
  pages={216--233},
  year={2024},
  organization={Springer}
}

@article{li2023seed,
  title={Seed-bench: Benchmarking multimodal llms with generative comprehension},
  author={Li, Bohao and Wang, Rui and Wang, Guangzhi and Ge, Yuying and Ge, Yixiao and Shan, Ying},
  journal={arXiv preprint arXiv:2307.16125},
  year={2023}
}

@inproceedings{hudson2019gqa,
  title={Gqa: A new dataset for real-world visual reasoning and compositional question answering},
  author={Hudson, Drew A and Manning, Christopher D},
  booktitle={Proceedings of the IEEE/CVF conference on computer vision and pattern recognition},
  pages={6700--6709},
  year={2019}
}

@inproceedings{yue2024mmmu,
  title={Mmmu: A massive multi-discipline multimodal understanding and reasoning benchmark for expert agi},
  author={Yue, Xiang and Ni, Yuansheng and Zhang, Kai and Zheng, Tianyu and Liu, Ruoqi and Zhang, Ge and Stevens, Samuel and Jiang, Dongfu and Ren, Weiming and Sun, Yuxuan and others},
  booktitle={Proceedings of the IEEE/CVF Conference on Computer Vision and Pattern Recognition},
  pages={9556--9567},
  year={2024}
}

@article{yu2023mm,
  title={Mm-vet: Evaluating large multimodal models for integrated capabilities},
  author={Yu, Weihao and Yang, Zhengyuan and Li, Linjie and Wang, Jianfeng and Lin, Kevin and Liu, Zicheng and Wang, Xinchao and Wang, Lijuan},
  journal={arXiv preprint arXiv:2308.02490},
  year={2023}
}

@article{ghosh2023geneval,
  title={Geneval: An object-focused framework for evaluating text-to-image alignment},
  author={Ghosh, Dhruba and Hajishirzi, Hannaneh and Schmidt, Ludwig},
  journal={Advances in Neural Information Processing Systems},
  volume={36},
  pages={52132--52152},
  year={2023}
}

@article{hu2024ella,
  title={Ella: Equip diffusion models with llm for enhanced semantic alignment},
  author={Hu, Xiwei and Wang, Rui and Fang, Yixiao and Fu, Bin and Cheng, Pei and Yu, Gang},
  journal={arXiv preprint arXiv:2403.05135},
  year={2024}
}

@article{ye2025imgedit,
  title={Imgedit: A unified image editing dataset and benchmark},
  author={Ye, Yang and He, Xianyi and Li, Zongjian and Lin, Bin and Yuan, Shenghai and Yan, Zhiyuan and Hou, Bohan and Yuan, Li},
  journal={arXiv preprint arXiv:2505.20275},
  year={2025}
}

@article{zheng2022movq,
  title={Movq: Modulating quantized vectors for high-fidelity image generation},
  author={Zheng, Chuanxia and Vuong, Tung-Long and Cai, Jianfei and Phung, Dinh},
  journal={Advances in Neural Information Processing Systems},
  volume={35},
  pages={23412--23425},
  year={2022}
}

@article{xie2024show,
  title={Show-o: One single transformer to unify multimodal understanding and generation},
  author={Xie, Jinheng and Mao, Weijia and Bai, Zechen and Zhang, David Junhao and Wang, Weihao and Lin, Kevin Qinghong and Gu, Yuchao and Chen, Zhijie and Yang, Zhenheng and Shou, Mike Zheng},
  journal={arXiv preprint arXiv:2408.12528},
  year={2024}
}

@article{liao2025mogao,
  title={Mogao: An omni foundation model for interleaved multi-modal generation},
  author={Liao, Chao and Liu, Liyang and Wang, Xun and Luo, Zhengxiong and Zhang, Xinyu and Zhao, Wenliang and Wu, Jie and Li, Liang and Tian, Zhi and Huang, Weilin},
  journal={arXiv preprint arXiv:2505.05472},
  year={2025}
}

@article{wang2024emu3,
  title={Emu3: Next-token prediction is all you need},
  author={Wang, Xinlong and Zhang, Xiaosong and Luo, Zhengxiong and Sun, Quan and Cui, Yufeng and Wang, Jinsheng and Zhang, Fan and Wang, Yueze and Li, Zhen and Yu, Qiying and others},
  journal={arXiv preprint arXiv:2409.18869},
  year={2024}
}

@article{wang2025image,
  title={Image Editing with Diffusion Models: A Survey},
  author={Wang, Jia and Hu, Jie and Ma, Xiaoqi and Ma, Hanghang and Wei, Xiaoming and Wu, Enhua},
  journal={arXiv preprint arXiv:2504.13226},
  year={2025}
}

@inproceedings{guo2025mammoth,
  title={Mammoth-vl: Eliciting multimodal reasoning with instruction tuning at scale},
  author={Guo, Jiawei and Zheng, Tianyu and Li, Yizhi and Bai, Yuelin and Li, Bo and Wang, Yubo and Zhu, King and Neubig, Graham and Chen, Wenhu and Yue, Xiang},
  booktitle={Proceedings of the 63rd Annual Meeting of the Association for Computational Linguistics (Volume 1: Long Papers)},
  pages={13869--13920},
  year={2025}
}

@article{an2025llava,
  title={Llava-onevision-1.5: Fully open framework for democratized multimodal training},
  author={An, Xiang and Xie, Yin and Yang, Kaicheng and Zhang, Wenkang and Zhao, Xiuwei and Cheng, Zheng and Wang, Yirui and Xu, Songcen and Chen, Changrui and Wu, Chunsheng and others},
  journal={arXiv preprint arXiv:2509.23661},
  year={2025}
}

@article{chen2025opengpt,
  title={Opengpt-4o-image: A comprehensive dataset for advanced image generation and editing},
  author={Chen, Zhihong and Bai, Xuehai and Shi, Yang and Fu, Chaoyou and Zhang, Huanyu and Wang, Haotian and Sun, Xiaoyan and Zhang, Zhang and Wang, Liang and Zhang, Yuanxing and others},
  journal={arXiv preprint arXiv:2509.24900},
  year={2025}
}

@misc{wang2025gptimageedit15mmillionscalegptgeneratedimage,
      title={GPT-IMAGE-EDIT-1.5M: A Million-Scale, GPT-Generated Image Dataset}, 
      author={Yuhan Wang and Siwei Yang and Bingchen Zhao and Letian Zhang and Qing Liu and Yuyin Zhou and Cihang Xie},
      year={2025},
      eprint={2507.21033},
      archivePrefix={arXiv},
      primaryClass={cs.CV},
      url={https://arxiv.org/abs/2507.21033}, 
}

@article{zhuunveiling,
  title={Unveiling the Secret of AdaLN-Zero in Diffusion Transformer},
  author={Zhu, Jie and Ding, Mingyu and Duan, Boqiang and Wang, Leye and Wang, Jingdong}
}

@article{zhu2025auditing,
  title={Auditing Data Provenance in Real-world Text-to-Image Diffusion Models for Privacy and Copyright Protection},
  author={Zhu, Jie and Wang, Leye},
  journal={arXiv preprint arXiv:2506.11434},
  year={2025}
}

@article{li2025onecat,
  title={Onecat: Decoder-only auto-regressive model for unified understanding and generation},
  author={Li, Han and Peng, Xinyu and Wang, Yaoming and Peng, Zelin and Chen, Xin and Weng, Rongxiang and Wang, Jingang and Cai, Xunliang and Dai, Wenrui and Xiong, Hongkai},
  journal={arXiv preprint arXiv:2509.03498},
  year={2025}
}
\bibliographystyle{iclr2025_conference}

\end{document}